\documentclass{article}
\usepackage{ijcai23}

\usepackage{times}
\usepackage{soul}
\usepackage{url}
\usepackage[hidelinks]{hyperref}
\usepackage[utf8]{inputenc}
\usepackage[small]{caption}
\usepackage{graphicx}

\usepackage{amsmath,amsfonts,amsthm,bbm} 
\usepackage{dsfont}
\usepackage{booktabs}
\usepackage{algorithm}
\usepackage{algorithmic}
\usepackage{xcolor}
\usepackage[switch]{lineno}

\usepackage{pifont}
\usepackage{float}

\newtheorem{example}{Example}

\newtheorem{definition}{Definition}
\newcommand{\argmaxI}{\mathop{\mathrm{argmax}}\nolimits} 

\title{HireVAE: An Online and Adaptive Factor Model Based on Hierarchical and Regime-Switch VAE}

\author{
Zikai Wei$^1$
\and
Anyi Rao$^2$\and
Bo Dai$^3$\And
Dahua Lin$^{1,3}$
\affiliations
$^1$The Chinese University of Hong Kong\\
$^2$Stanford University\\
$^3$Shanghai AI Laboratory
}

\begin{document}

\maketitle

\begin{abstract}
    
    Factor model is a fundamental investment tool in quantitative investment, 
    which can be empowered by deep learning to become more flexible and efficient in practical complicated investing situations.
    However, it is still an open question to build a factor model that can conduct stock prediction in an online and adaptive setting,
    where the model can adapt itself to match the current market \textbf{re}gime identified based on only point-in-time market information.
    To tackle this problem, we propose the first deep learning based online and adaptive factor model, \textbf{HireVAE},
    at the core of which is a \textbf{hi}erarchical latent space 
    that embeds the underlying relationship between the market situation and stock-wise latent factors,
    so that HireVAE can effectively estimate useful latent factors given only historical market information and subsequently predict accurate stock returns.
    Across four commonly used real stock market benchmarks,
    the proposed HireVAE demonstrate superior performance 
    in terms of active returns over previous methods, verifying the potential of such online and adaptive factor model.
\end{abstract}

\section{Introduction}


Originating in asset pricing, factor model is the foundation of factor-based investing, which simplifies the high-dimensional characteristics in quantitative investment via an experimental way to model how factors drive returns and risks.
This can significantly help institutional investors construct diverse and customized portfolios.
But, a better-developed capital market brings out more challenges in achieving excess returns in the market using traditional methods due to more similar investing actions and overlapping trading positions. 
Although 
the recent nonlinear data-driven methods \cite{duan2022factorvae,wei2022factor,gu2021autoencoder} enrich 
the conventional tool library built largely from linear methods \cite{ng1992multi,fama2020comparing},
they still face serious issues in handling large, abrupt, and sustained market changes and some factors that worked well in the previous market situation become less effective in the new situation.
Therefore, it is of great importance to model the current market regime and develop an adaptive factor model associated with different market situations.

Both practitioners and academics make many attempts to model the regime change in markets using statistical learning \cite{guidolin2007asset}, machine learning \cite{uysal2021machine}, or hybrid methods \cite{akioyamen2020hybrid} from different perspectives of the market.
Existing data-driven approaches view regime switching as a clustering problem and represent each regime as a cluster, subdividing the market regime based on the information absorbed from the entire training data. 
However, this setting is not suitable for learning a regime-switching model intended for later use in investment practice. Practice requires an online learning setting that all data used for learning should be point-in-time data, meaning that there is no future information leakage at any timestamp in the training data set. 

In this paper, we thus propose a novel end-to-end neural factor model, referred to as HireVAE, that offers an online and adaptive regime-switching capability,
targeting two critical questions:
\noindent\textbf{Q1:} How to keep the regime cluster labels changing consistently and centers of cluster shifting smoothly during training?
\noindent\textbf{Q2:} How to adaptive learn regime-oriented factor model under different regimes?

To tackle \textbf{Q1}, we develop a module for identifying regimes, 
in which it learns a set of dynamic distributions associated with different regime clusters. To maintain this regime identifier as a stable clustering module, we design a linear stabilization algorithm to supervise the identifier learn a consistent clustering that smoothly alternates with a market latent variable.
Specifically,
a market encoder, which is learned in a prior-posterior learning framework, will encode the market information into a market latent variable.
This framework can help the market encoder learn the hidden logic between the market information and the average future stock returns in the whole market,
which will serve as prior knowledge for the market encoder in predicting future stock returns given only current market information.
The market latent variable plays two roles in HireVAE: 1) at first it is used to estimate the means and variances of $K$ Gaussian distributions, representing $K$ regimes. 
2) Subsequently, it will also be used to predict $K$ regime indicators measuring the likelihood of each regime being the most suitable one,
from which the one with the largest likelihood can thus be identified.
It's worth noting that we will sort $K$ estimated regimes by sorting their means in descending order,
as pointed out by \cite{botte2021machine},
where returns and risk-adjusted measures are linearly correlated across market regimes, so that we can partition the whole market situation into market regions following a consecutive order.
With the help of the market latent variable,
we can then adaptively select the most suitable regime given current market information.
However, one question remains: How to smoothly update the regime centers and make a consistent regime assignment?
To ensure consistency of cluster assignment during training, the cluster ID is always reordered in the descending order. In this way, the market latent variable can always be associated with a consistent regime cluster and the additional burden of checking for a ``wrong assignment" is eliminated.
Moreover, to smoothly update the regime clusters, we use a dynamically weighted moving average algorithm to update the means and standard deviations of the clusters, giving more weight to the historical clusters.
To cope with \textbf{Q2}, we develop an encoder with a hierarchical latent space of market and stock latent variables and a set of regime-based stock decoders, each of which recovers or predicts stock returns based on regime, market, and stock latent factors. With this linear stabilization algorithm for learning market regimes, the end-to-end training framework can adpatively learn different market conditions as well as the corresponding regime-specific factor model for stock prediction. In a later section, we will show the advantages of this algorithm for regime switching with adaptive learning 
compared to rule-based regime switching, e.g., categorizing market regimes with respect to different volatility levels.

\begin{figure}[t]
\centering
\includegraphics[width=1.0\linewidth]{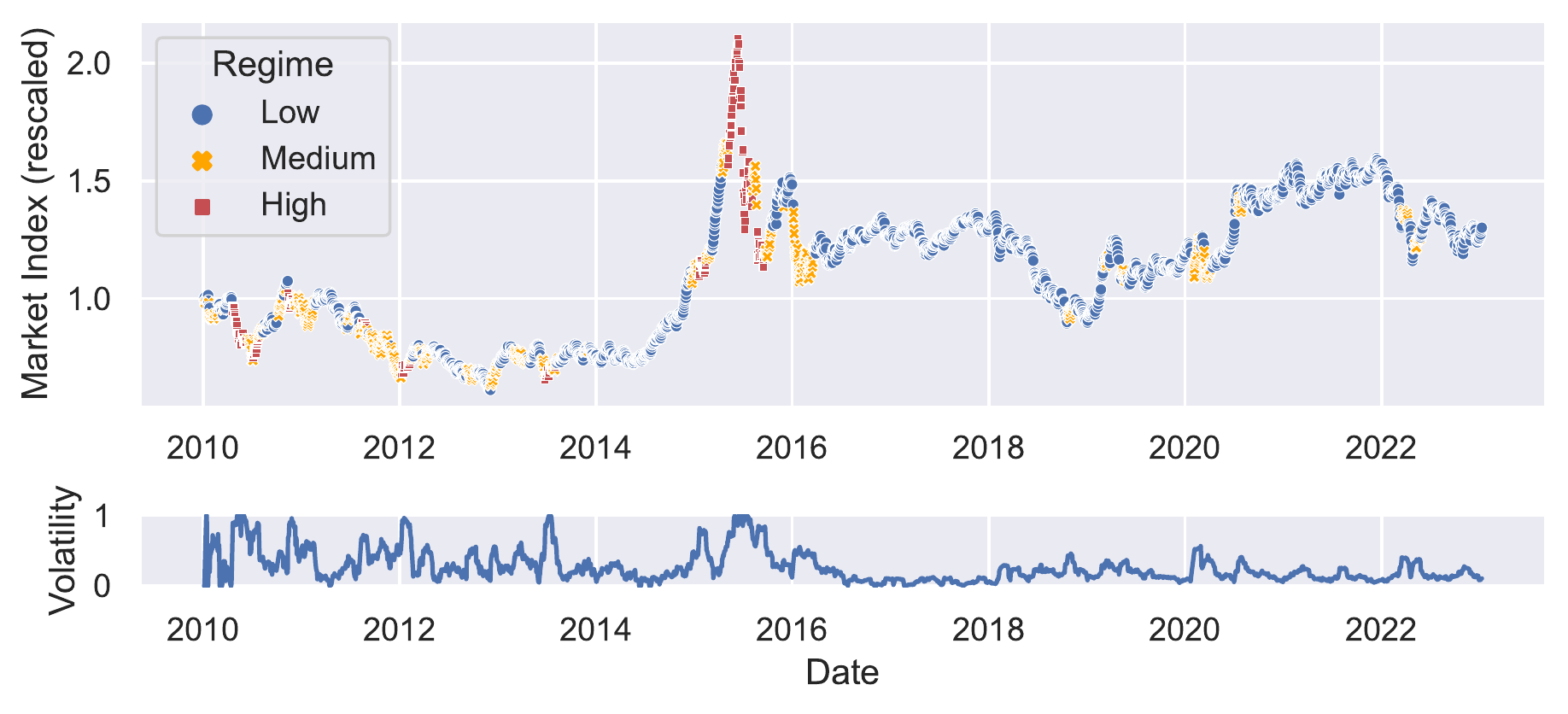}
\caption{Identifying market regimes can help make better investment decisions, and there is an example that identifies regime changes based on volatility, where volatility acts as an experience-and-knowledge based indicator that is widely used in practice.} \label{market_sigma}
\end{figure}

\begin{example}[Determining regimes helps better decision making]
 Figure \ref{market_sigma} shows the structural changes of the CSI All Share Index, composed of all stocks in China stock market, in terms of volatility level.
 When volatility is high, there is usually a change from a bullish market to a bearish market. In contrast, medium volatility usually occurs during the period when the market is changing trend at a more steady pace. The market begins a long bull run when the low volatility prevails most of the time.
\end{example}
The contribution of our paper are as follows: 1) To the best of our knowledge, we are the first to propose regime switching with online regime learning in an end-to-end training framework. 
    2) We provide an adaptive learning framework for training a regime switch factor model  with a hierarchical latent space that incorporates global observation of stock market and stock-oriented latent factors.  The decoders are also online trained based on different dynamic market conditions, each of which takes advantage of regime, market, and stock information in the following downstream tasks.
    3) We develop a linear stabilization algorithm that helps the regime identification process to learn a more consistent and smoother clustering deep learning technique.
\section{Related Work}
\paragraph{Factor Models.} The factor model are traditionally regarded as a pricing model in academia \cite{fama1992cross,fama2020comparing}, while practitioners use its anomalies as factors to capture excess returns from mispricing \cite{levin1995stock}.
In recent years, stock forecasting studies based on machine learning methods are emerging.
Machine learning based stock prediction methods are closely related to factor models and can be categorized as factor mining and factor composition. The former learns a new logical factor from new data sources or perspective, while the latter finds a better way to compose the existing factors, which can be considered as a machine learning based factor model.
In terms of factor mining, existing works on stock prediction study trading patterns \cite{wang2022adaptive,ding2020hierarchical,zhang2017stock}, investment reviews \cite{wangheterogeneous}, overnight patterns \cite{li2021modeling}, and the temporal relationship of stocks \cite{wang2021hierarchical}.
As for factor composition as nonlinear factor models, existing works learn optimal latent factors via a variational autoencoder (FactorVAE) \cite{duan2022factorvae}, a deep multifactor model (DMFM), which builds on hierarchical stock graphs \cite{wei2022factor}, and a deep risk factor model that fuses a set of uncorrelated risk factors from the original style factors \cite{lin2021deep}. 
Unlike existing methods, our model aims to learn an adaptive factor model that can recognize the current market regime and use a learnable and more appropriate factor composition for that market regime, which can enable a market-driven investment decision.

\paragraph{Regime Switching} \cite{hamilton1989new} provide a statistical modeling framework to address the problem of regime switching in various circumstances, such as regime switching in interest rates \cite{ang2002regime}, asset allocation \cite{guidolin2007asset}, and financial markets\cite{zhu2022regime}. In the cryptocurrency market, \cite{zhu2022regime} shows that regime switching methods perform better than methods without regime switching. Machine learning based methods open another main direction in modeling regime switching, such as Hidden Markov Model \cite{jrfm13120311} and Gaussian Mixed Models \cite{botte2021machine}. As a hybrid method, \cite{akioyamen2020hybrid} applies principal component analysis and $k$-means clustering to identify regimes in financial markets. Deep learning based regime switching models of energy commodity prices \cite{mari2022deep}. Unlike the existing regime switching method, in investment practice, we not only need to use an online learning algorithm to identify market regimes, but also need to learn an adaptive factor model based on the determination of the market regime.

\paragraph{Variational Autoencoder.} 
Variational Autoencoder (VAE) \cite{kingma2013auto}, one of the major families of deep generative models, provides a prior-posterior framework that encodes the observation as prior knowledge in a latent space, and later \cite{vahdat2020nvae} introuduce a hierarchical pipeline into VAE.
\cite{chung2015recurrent} is the first to extend VAE to sequential modeling. Then, the variational methods are extended to neural machine translation \cite{su2018variational} and trajectory prediction. In the area of financial applications, VAE includes topics such as stochastic volatility models \cite{luo2018neural} and factor models. The latter VAE family includes applications in factor mining based on social media in the context of stocks \cite{xu2018stock}, applications in factor composition where \cite{duan2022factorvae} aims to learn representative latent factors from existing technical factors, and \cite{gu2021autoencoder} aims to learn a pricing model.
\section{Preliminaries}
In this section, we first define the stock prediction problem and then introduce the problem setting from the perspective of learning a data-driven method.
\begin{definition}[Stock prediction]
    In general, stock prediction problem can be defined as learning 
    a mapping $f\left( \cdot ; \Theta \right)$
    from the past information set $\mathcal{F}_{<t}$ to future stock return $ \mathbf{y}_{t+\Delta_t}$, where $\Delta_t$ is the prediction length, 
     $\hat{\mathbf{y} }_{t+\Delta_t}=\mathbb{E}\left( \mathbf{y}_{t+\Delta_t} | \mathcal{F}_{<t}\right)$
    is its corresponding expected future return, and
    $\Theta$ is the model parameter of data-driven method $f\left( \cdot\right)$. 
\end{definition}
In our setting, we have a set of samples labeled with time stamps sorted in chronological order: $\mathcal{D} = \{\mathcal{D}_{t_0}, \dots, \mathcal{D}_{t_{n}} \}$, where $n$ is the total number of trading days covered by the dataset. 
For a single data sample $\mathcal{D}_{t} = \{\mathbf{X}_{t},
\mathbf{M}_{t} \}$,  $\mathbf{X}_{t} \in \mathbb{R}^{T \times N \times C}$ and $\mathbf{M}_{t} \in \mathbb{R}^{T \times C_m}$ are sequential features of stocks and the market, where $T$, $N$ and $C_{ (\cdot)}$ represent the length of the historical sequence, the number of stocks and the number of features. Moreover, the overall pattern of the market (global observation) consists of information from $d$ modalities, based on which it can be segmented as $\mathbf{M}_{t} = [\mathbf{M}_{t}^1, \dots , \mathbf{M}_{t}^d]$. The above are general input dataset.

Besides, the corresponding target set of $\mathcal{D}$ is $\mathcal{Y} = \{\mathcal{Y}_{t_0}, \dots, \mathcal{Y}_{t_{n}} \}$, where $\mathcal{Y}_{t} = \{ \mathbf{y}_{t+\Delta t} \}$, $\Delta t$ is the length of prediction horizon, and $\mathbf{y}_{t+\Delta t} \in \mathbb{R}^{N}$ is the future cross-section returns of $N$ stocks in the market over the time period from $t$ to $t+\Delta t$.   

Our target is to learn a optimal $\Theta^*$ of a data driven model $f\left(\cdot; \Theta \right)$ with learnable parameter $\Theta$ such that $\hat{\mathcal{Y}} \leftarrow f\left(\mathcal{D}; \Theta \right)$.
\section{Methodology}
\begin{figure}[t]
\centering
\includegraphics[width=\linewidth]{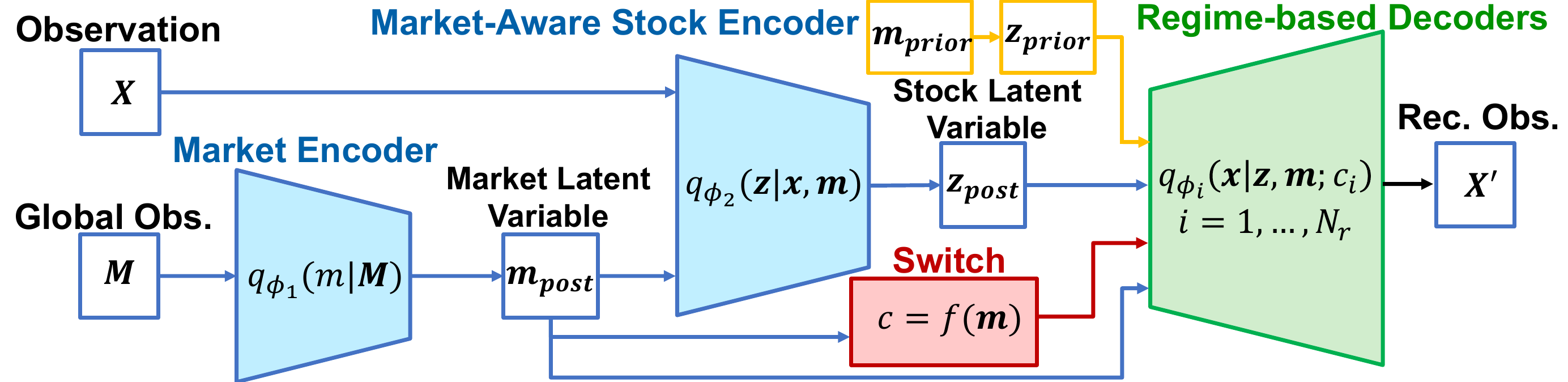}
\caption{The brief architecture of HireVAE.} \label{fig1_brief}
\end{figure}

In this section, we introduce HireVAE to identify the market regime with online learning in conjunction with a linear stabilization clustering algorithm and learn an adaptive factor model to extract the latent logic behind the market and stocks under different regimes. The brief framework of HireVAE is given in Figure \ref{fig1_brief}.

The architecture of our model is derived from a VAE encoder-decoder structure. In contrast to the classical pipeline, we develop a hierarchical encoder with a two-level latent space representing a market-stock hierarchy, and couple it with a set of regime-specific decoders that reconstruct or predict stock returns based on different market situations. First, the market encoder extracts market latent factors from multimodal features. Then, the stock encoder plays the role of a conditional encoder that extracts stock latent factors based on stock features and market latent factors.
This hierarchical structure can help the stock latent factor better recognize the market situation.
In the middle of this encoder-decoder architecture, a regime-switching module acts as a mediator to match an appropriate regime-specific decoder to the latent market situation.

Specifically, this architecture consists of four parts: an explanation of the feature extraction procedure from sequential data, an overview of the hierarchical encoder-decoder structure, a regime switching learning algorithm, and the learning objectives in prior-posterior learning.

\subsection{Extracting Features from Sequential Data} \label{sec:features}

We establish two inherent feature extractors, \emph{i.e.}, stock-specific and market-specific extractors. 
\paragraph{Stock Features.} The stock feature extractor $\phi^{feat}_s$, extracts features $\mathbf{e}_s$ from the historical stock characteristics: $\mathbf{e}_s = \phi^{feat}_s(\mathbf{X})$, where $\mathbf{X} \in \mathbb{R}^{T \times N \times C}$ and $\mathbf{e}_s \in \mathbb{R}^{N \times H}$ represents the extracted stock features with a hidden size $H$. Specifically, $\mathbf{e}_s$ is the latest hidden state obtained by a gated recurrent united which summarizes all temporal information.

\paragraph{Market Features.} 
For sequential market data $\mathbf{M}_t \in \mathbb{R}^{T\times C_m}$ consists of information with $d$ multimodal features,
we introduce a set of extractors $\phi^{feat}_{m,i}(\cdot)$ associated with different modalities $\mathbf{M}^i_t \in \mathbb{R}^{T\times C_i}$, $i=1,\dots,d$, and $\sum_{i=1}^d {C^i_t} = C_m$. We use $d!$ cross attention modules \cite{hou2019cross} to learn pairwise features from each modality pair. Later, we extract the market features $\mathbf{v}$ from all these pairwise features by concatenation. In our setting, a market can be represented by an index composed of all the stocks in the market. The different modalities are the market momentums (defined as the cumulative return over a period of time), the levels of relative trading volume, and the market volatilities over different time frames.

\subsection{Hierarchical Encoder-Decoder Structure}
To help the model better identify the market situation and better interpret future stock returns in different market situations, we design the latent space in a hierarchical structure that represents latent factors from the market and stocks. 
A pair of encoders is designed to learn latent variables from market and stocks based on the market and stock features. 

\subsubsection{Posterior Structure in Training Phase} 
In the training phase, the market encoder generates a market latent variable, based on which and in conjunction with both future stock returns and historical stock features a market-aware stock encoder can learn stock latent features. After the current market regime is identified by a regime switching algorithm, a regime-specific decoder can reconstruct future stock returns. 
\paragraph{Market Encoder.} The market encoder $\phi_{enc}^{m}$ extracts posterior market latent factor $\mathbf{m}$ 
from the extracted market information $\mathbf{v}$,
and future average return of all stocks $\bar{\mathbf{y}}$:
\begin{align*}
    &[\boldsymbol{\mu}^m_{post}, \boldsymbol{\sigma}^m_{post}] = \phi_{enc}^{m}(\mathbf{v}, \bar{\mathbf{y}}),\\
    &\mathbf{m}  \sim \mathcal{N}\left(\boldsymbol{\mu}^m_{post}, \text{diag}(\boldsymbol{\sigma}^m_{post})\right),
\end{align*}
where $\mathbf{m} \in \mathbb{R}^{H_m}$ is the market latent factor following a Gaussian distribution, parameterized by mean $\boldsymbol{\mu}^m_{post}$ and standard deviation $\boldsymbol{\sigma}^m_{post}$ and ${H_m}$ is the market hidden size.

\paragraph{Market-Aware Stock Encoder.} The stock encoder $\phi_{enc}^{s}$ extracts posterior stock latent factors $\mathbf{z}$ from the market latent factor $\mathbf{m}$, stock features $\mathbf{e}_s$, and future stock returns $\mathbf{y}$:
\begin{align*}
    &[\boldsymbol{\mu}^s_{post}, \boldsymbol{\sigma}^s_{post}] = \phi_{enc}^{s}\left(\mathbf{v}, \mathbf{e}_s, \mathbf{y}\right),\\
    &\mathbf{z}  \sim \mathcal{N}\left(\boldsymbol{\mu}^s_{post}, \text{diag}(\boldsymbol{\sigma}^s_{post})\right),
\end{align*}
where $\mathbf{z} \in \mathbb{R}^{H_s}$ is the stock latent factors following a Gaussian distribution, parameterized by mean $\boldsymbol{\mu}^s_{post}$ and standard deviation $\boldsymbol{\sigma}^s_{post}$ and ${H_s}$ is the stock hidden size.

\paragraph{Regime-Switch Factor Decoder.} 
We develop a learning algorithm to distinguish the current market situation from different regimes in the latter part of the paper (Section \ref{regime-learning}). 
To learn an adaptive factor model for different market situations, we design a regime-switch factor decoder consisting of $N_k$ sub-decoders that can reconstruct future stock returns $\hat{\textbf{y}}$ based on stock latent factors $\mathbf{z}$ and market latent factors $\mathbf{m}$:
\begin{align*} \hat{\mathbf{y}} = \phi_{dec}\left( \mathbf{z}, \mathbf{m}, \mathbf{e}_s; c \right),
\end{align*}
where $c \in \{1,\dots,N_k\}$ denotes a market regime. More precisely, $c = f(\mathbf{m})$ where $f(\cdot)$ is a switching function (algorithm) that can identify the regime based on the market latent factors $\mathbf{m}$.

\subsubsection{Prior Structure in Prediction Phase}
First, the market prior generates market latent factors $\mathbf{m}_0$ based on the extracted market information $\mathbf{v}$ from various modalities. Later, the stock prior generates stock latent factors $\mathbf{z}_0$ based on the market prior variables $\mathbf{m}_0$ and stock characteristics $\mathbf{e}_s$. The decoder predicts future returns based on the market-stock priors and the regime $c$ determined by the switch function. Specifically, 
\begin{align*}
    &[\boldsymbol{\mu}^m_{0}, \boldsymbol{\sigma}^m_{0}] = \phi_{prior}^{m}(\mathbf{v}),\\
    &\mathbf{m}_0  \sim \mathcal{N}\left(\boldsymbol{\mu}^m_{0}, \text{diag}(\boldsymbol{\sigma}^m_{0})\right), \\
    &[\boldsymbol{\mu}^s_{0}, \boldsymbol{\sigma}^s_{0}] = \phi_{prior}^{s}\left(\mathbf{m}_0, \mathbf{e}_s\right),\\
    &\mathbf{z}_0  \sim \mathcal{N}\left(\boldsymbol{\mu}^s_{0}, \text{diag}(\boldsymbol{\sigma}^s_{0})\right),\\
    &\hat{\mathbf{y}} = \phi_{dec}\left( \mathbf{z}_0, \mathbf{m}_0, \mathbf{e}_s; c \right) \text{ with } c = f(\mathbf{m}_0),
\end{align*}
where $\mathbf{m}_0 \in \mathbb{R}^{H_m}$ follows a Gaussian distribution with parameters $\boldsymbol{\mu}^m_{0}$ and $\boldsymbol{\sigma}^m_{0}$, and $\mathbf{z}_0 \in \mathbb{R}^{H_s}$ follows a Gaussian distribution with parameters $\boldsymbol{\mu}^s_{0}$ and $\boldsymbol{\sigma}^s_{0}$.

\begin{algorithm}[tb]
    \caption{Linear stabilization clustering algorithm}  
    \label{alg:algorithm}
    \textbf{Input}: Training data $\mathcal{D}^{train}= \{\mathcal{D}^{batch}_{1}, \dots, \mathcal{D}^{batch}_{N_b}\}$ \emph{s.t.} all data items are sorted in chronological order \\
    \textbf{Parameter}: Model parameters $\Phi$, regime distribution parameters $\boldsymbol{\mu}_r \in \mathbb{R}^{N_r}$ and $\boldsymbol{\sigma}_r \in \mathbb{R}^{N_r}_{+}$ \\
    \textbf{Output}: Regime $\mathbf{c}_i, i=1,\dots, N_b$
    \begin{algorithmic}[1] 
        \STATE Randomly initialize $\boldsymbol{\mu}_r$ and $\boldsymbol{\sigma}_r$ with $\boldsymbol{\sigma}_r \geq 0$ 
        \WHILE{$ i \leq N_b$ }
        \STATE Projection to 1D score space: $\mathbf{s}_i \leftarrow  \texttt{Score}\left(\mathcal{D}^{batch}_{i}\right)$
        \STATE Pred. cluster center: $\boldsymbol{\mu}_i, \boldsymbol{\sigma}_i \leftarrow  \texttt{DistShift}\left(\mathcal{D}^{batch}_{i}\right)$
        \STATE Sort cluster center: $\boldsymbol{\mu}_i^*, \boldsymbol{\sigma}_i^* \leftarrow  \texttt{Sort}\left(\boldsymbol{\mu}_i, \boldsymbol{\sigma}_i\right)$
        \STATE Update cluster center: \\$~~~~~~\boldsymbol{\mu}_r \leftarrow \beta \boldsymbol{\mu}_r + (1-\beta) \boldsymbol{\mu}_i^*$
        \\ $~~~~~~\boldsymbol{\sigma}_r \leftarrow \beta \boldsymbol{\sigma}_r + (1-\beta) \boldsymbol{\sigma}_i^*$
        \STATE Calc. regime prob.: $\boldsymbol{p}_i \leftarrow \mathcal{N}\left(\mathbf{s}_i; \boldsymbol{\mu}_r, \text{diag}(\boldsymbol{\sigma}_r^2)\right)$
        \STATE Pred. regime: $\mathbf{c}_i \leftarrow  \texttt{ArgMax}\left(\mathbf{p}_i\right)$
        \ENDWHILE
        \STATE \textbf{return} regime $\mathbf{c}_i, i=1,\dots, N_b$
    \end{algorithmic}
\end{algorithm}

\subsection{Regime Switching with Online Learning} \label{regime-learning} 
In order to learn an adaptive factor model that can handle various regimes, 
the first step is to identify these regimes based on point-in-time (PIT) market information.
To achieve consistent and smooth market clustering based on a market latent variable $\mathbf{m} \in \mathbb{R}^{H_m}$ extracted from the mixed market features $\mathbf{v} \in \mathbb{R}^{H_c}$, we develop an online learning clustering algorithm with linear stabilization summarized in Algorithm \ref{alg:algorithm}. First, we project the market latent variable onto a 1D measure space, and consider its relative score in this 1D space as a random variable from a market regime distribution. Then, we compute the log-likelihood of this score \emph{w.r.t} the distributions of different market regimes and choose the regime with largest log-likelihood as the predicted regime.

\paragraph{Projection to 1D Space.} A project layer $\phi^{proj}$ projects the market latent variable $\mathbf{m} \in \mathbb{R}^{H_m}$ onto a 1D regime measure space. We define $s = \phi_{proj}(\mathbf{m})$, $s\in \mathbb{R}$, as the ``score" of the current market situation and it plays a role as a 1D measure of the overall market status. This market score $s$ is considered as a given sample in the sample spaces of the different regimes. Later, we can measure how much more likely this market score $s$ belongs to a given market regime.

\paragraph{Determination of Market Regimes.} Denote the number of market regimes as $N_r$. We assume that $\mathbf{r}$ is a Gaussian random vector given by $\mathbf{r} \sim \mathcal{N}(\boldsymbol{\mu}_r, \text{diag}(\boldsymbol{\sigma}_r))$, where $\boldsymbol{\mu}_r \in \mathbb{R}^{N_r}$ and standard deviation $\boldsymbol{\sigma}_r \in \mathbb{R}^{N_r}_{+}$ are the representatives of regime centers and regime deviations \emph{w.r.t} ${N_r}$ regime clusters. To keep each regime as different as possible, we use Kullback-Leibler divergence (KL) to measure the difference between two distributions and define the loss function in terms of market regimes as 
\begin{equation*}
    L_{reg} = - \sum_{\substack{i=1,\dots,N_r\\j= i+1,\dots, N_r}}\text{KL}\left( \mathcal{N} \left( {\mu}_i, {\sigma}_i^2\right)||\mathcal{N} \left( {\mu}_j,{\sigma} _j^2 \right)\right),
\end{equation*}
where ${\mu}_i$ and ${\sigma}_i$ are the $i$-th element of the mean vector $\boldsymbol{\mu}_r$ and the deviation vector $\boldsymbol{\sigma}_r$, respectively.

\paragraph{Regime Learner.} To determine the centers of each regime, the regime learner $\phi_{reg}$ predicts the parameters of the market regime distribution, \emph{i.e.} $[\boldsymbol{\mu}_r, \boldsymbol{\sigma}_r] = \phi_{reg}\left( \mathbf{m} \right)$. Specifically,
\begin{align*}
    \boldsymbol{\mu}_r &= \mathbf{w}_{\mu,r} \mathbf{v} + \mathbf{b}_{\mu,r}, \\
    \boldsymbol{\sigma}_r & = \text{Softplus} \left(\mathbf{w}_{\sigma,r} \mathbf{v} + \mathbf{b}_{\sigma,r} \right).
\end{align*}
To keep the shift of regimes consistent and smooth in 1D measure space, a sorting function is applied to the mean values of all regime centers $\boldsymbol{\mu}_r \in \mathbb{R}^{N_r}$ in descending order and the corresponding deviation terms are reordered \emph{w.r.t.} the new order of sorted mean values. Formally, 
\begin{align*}
    [\boldsymbol{\mu}_r^{*}, \boldsymbol{I}_r^{*}] & = \text{Sort} \left(\boldsymbol{\mu}_r \right), \\
    \boldsymbol{\sigma}_r^{*} &= \boldsymbol{\sigma}_r[ \boldsymbol{I}_r^{*}],
\end{align*}
where $\boldsymbol{\mu}_r^{*}$ is a reordered mean vector, $\boldsymbol{I}_r^{*}$ is its corresponding reordered index \emph{w.r.t.} its original index, and $\boldsymbol{\sigma}_r^{*}$ is a reordered deviation vector \emph{w.r.t.} this reordered index $\boldsymbol{I}_r^{*}$. These reordered mean and deviation vectors obtained in each batch are later used to update each regime center ($\boldsymbol{\mu}_r$ and $\boldsymbol{\sigma}_r$) with a weighted moving average, allowing the regime center to be updated gradually.

\paragraph{Regime Prediction.} We assume that the market score $s$ is a random variable generated from a market regime distribution defined as a Gaussian distribution. For $N_k$ regimes, we have $N_k$ market regime distributions, for each of which we can calculate the log-likelihood of the current market score $s$:
\begin{align*}
    l\left(s; {\mu}_r^{i},\sigma_r^{i}\right) &= -\ln({\sigma}_r^{i}) - \frac{1}{2} \ln(2  \pi) - \frac{1}{2}\left(\frac{{({s} -{\mu}_r^{i})}^2}{{\sigma}_r^{i}}\right), \\
\end{align*}
where $i=1,\dots, N_k$.
The regime is predicted to be the regime with the maximum log-likelihood among all regimes: 
\begin{align*}
    c = \argmaxI_{j}  [l\left(s; {\mu}_r^{i},\sigma_r^{i}\right), i=1,\dots,N_k],
\end{align*}
where $c$ indicates the predicted market regime.


\subsection{Learning Objective}
Our learning objective consists of three parts, 
where the first part is to learn a better future stock return reconstruction, the second part is to learn latent market features and latent stock features in a hierarchical structure,
and the third part is to learn a more differentiated regime clusters. Thus, the overall loss function of our HireVAE is as follows,
\begin{align*}
    L_{rec} =~& - \sum_{r=1}^{N_r} \mathds{1}_{f(\mathbf{m})}(r)  \ln{P_{\phi_{dec}}\left(\hat{\mathbf{y}} = {\mathbf{y}} | \mathbf{X}, \mathbf{M}, \mathbf{z}, \mathbf{m}; r\right)}\\
    L_{hier} = ~&\text{KL}\left( P_{\phi_{enc}}\left(\mathbf{m}|\mathbf{M}, \bar{\mathbf{y}}\right)|| P_{\phi_{prior}}\left( \mathbf{m}|\mathbf{M},\bar{\mathbf{y}}\right) \right)+\\
     &\text{KL}\left( P_{\phi_{enc}}\left(\mathbf{z}|\mathbf{X},\mathbf{M},\mathbf{m},\mathbf{y}\right)|| P_{\phi_{prior}}\left( \mathbf{z}|\mathbf{X},\mathbf{M},\mathbf{m}\right) \right)\\
    L_{overall}=~& L_{rec} + L_{hier} + L_{reg}. 
\end{align*}

\section{Experiments}
\begin{table}[t]
    \centering
    \begin{tabular}{lrrr}
        \toprule
        Method  & IC & Rank IC & Rank ICIR \\
        \midrule
        Linear       & 0.030          & 0.031       & 0.322   \\
        GRU      & 0.046          & 0.050        & 0.484   \\
        MLP      & 0.053          & 0.051       & 0.537   \\
        Trans    & 0.050          & 0.040       & 0.253   \\
        GAT      & 0.029          & 0.032       & 0.466   \\
        IGAT     & 0.012          & 0.018       & 0.210  \\
        DMFM     & 0.014          & 0.015       & 0.480  \\
        VAE      & 0.049          & 0.059       & 0.539 \\
        CVAE     & 0.053        & 0.063       & 0.628   \\
        HiReVAE  & \textbf{0.058}         & \textbf{0.066}       & \textbf{0.734}   \\
        \bottomrule
    \end{tabular}
    \caption{Evaluation the performance of various methods in terms of their predictive power.}
    \label{tab:pred-power}
\end{table}
\begin{table*}[t]
  \centering
  \begin{tabular}{lrrrrrrrrrr}
    \toprule 
    &\multicolumn{2}{c}{CSI 300}  &\multicolumn{2}{c}{CSI 500}   &\multicolumn{2}{c}{CSI 1000}   &\multicolumn{2}{c}{CNI 2000}  &\multicolumn{2}{c}{CSI ALL}               \\
    Method     & AR$\uparrow$ &IR$\uparrow$ & AR$\uparrow$ &IR$\uparrow$& AR$\uparrow$ &IR$\uparrow$ & AR$\uparrow$ &IR$\uparrow$ & AR$\uparrow$ &IR$\uparrow$\\
    \midrule
    Linear   & 11.94  &0.30 &8.81 & 0.50  & 6.98  &0.65 &5.04 & 0.49&5.82 & 0.89  \\
    GRU   & 19.40  &0.60 &15.04 & 0.80  & 12.89  &0.96 &11.27 & 0.84&13.25 & 2.20 \\
    MLP    & 23.04  &0.78 &20.20 & 1.10  & 18.12  &1.22 &16.87 & \underline{1.09}&19.25 & 3.53 \\
    Trans  & 7.21  &0.18 &4.63 & 0.28  & 3.03  &0.41 &0.68 & 0.24& 0.22 & 0.27  \\
    GAT     & 5.77  &0.14 &5.65 & 0.19  & 4.33  &0.29 &2.30 & 0.16& 2.08 & 0.31 \\
    IGAT     & 12.67  &0.41 &10.32 & 0.61  & 8.47  &0.69  &6.85 & 0.54&8.43 & 2.04 \\
    DMFM    & 3.16  &0.05 &2.13 & 0.20  & 0.78  &0.35  &1.50 & 0.18& 2.24&0.32  \\
    VAE     & 26.49  &0.82 &25.04 & 1.11  & 23.32 &1.23 &19.05 & 1.02&23.72 & 4.38  \\
    CVAE     & \underline{27.78}  &\textbf{0.91} &\underline{25.84} & \underline{1.24}  & \underline{24.00}  &\textbf{1.36} &\underline{22.76} & 0.99&\underline{24.91} & \underline{5.46} \\
    \textbf{HiReVAE}     & \textbf{30.89}  &\underline{0.89} &\textbf{29.05} & \underline{1.23}  &\textbf{27.14}  &\underline{1.35} &\textbf{25.94} & \textbf{1.18}& \textbf{28.02} & \textbf{5.92} \\

    \bottomrule
  \end{tabular}
  \caption{The performances on portfolio construction over the period from mid-2015 to mid-2022 (\textbf{best}/\underline{2nd best}). 
  \label{tab:portfolio}
  }
\end{table*}
In this section, we evaluate the proposed HireVAE on real stock market data, choosing one of the largest emerging markets as representative, and demonstrate the effectiveness of our model through various experiments. These experiments are designed to investigate the following research questions.

\noindent\textbf{Exp 1:} Does our method outperform the state-of-the-art methods in terms of factor investing? 

\noindent\textbf{Exp 2:} Is the improvement in effectiveness mainly due to the additional market information?

\noindent\textbf{Exp 3:} How does regime switch help adaptive learning in factor models?

\noindent\textbf{Exp 4:} Is the hierarchical structure necessary in our design?


\subsection{Experiment Settings}\label{exp-setting}
\paragraph{Dataset Construction.} We conduct the experiment over the entire China stock market, with price-volume and fundamental data obtained from the WIND database, and all raw data processed on a point-in-time basis. The stock pool is constructed based on all on-listing stocks, except for stocks that receive special treatments (ST) and those that have been on the market for less than three months. When a company is labeled as ST, which means it suffer losses for two or more consecutive years. We follow the same paradigm as in \cite{wei2022factor} to construct stock features and select 58 features that have a good coverage over the entire stock market. 
The length of the sequential data and the prediction horizon are $T=20$ and $\Delta t =20$, respectively. The stock returns are calculated on a basis of volume weighted average prices.
We choose the dataset and data split settings as in \cite{wei2022factor} instead of \cite{duan2022factorvae} because the former better covers the period from 2010 to mid-2022 and has a longer test set from mid-2015 to mid-2022 than the latter from 2019 to 2020. 
There are 14 groups of training, validation, and test datasets. 
To prevent information leakage, we delete the data items from the validation dataset when the future stock returns in the validation dataset are covered by the time period of the test dataset.

\paragraph{Baselines.} To allow a fair comparison, we provide the same sequential data to all methods. We compare our HiReVAE to the factor models as follows:
\ding{172} \textbf{Linear} is a linear factor model derived from \cite{fama1992cross}, with average pooling on the sequential dimension. \ding{173} \textbf{GRU} \cite{chung2014empirical} is a factor model in which a GRU extracts the sequential features and, in conjunction with a linear layer, predicts future returns. 
\ding{174} \textbf{Trans} \cite{ding2020hierarchical} is a factor model based on Transformer.
\ding{175} \textbf{MLP} \cite{levin1995stock} is the first non-linear factor model with multilayer perceptrons, where we equip it with a GRU to extract time-series features.
\ding{176} \textbf{GAT} \cite{veličković2018graph} is a graph attention work with a fully connected stock graph.
\ding{177} \textbf{IGAT} \cite{wei2022factor} is a GAT, based on a stock-industry graph where there is an edge between two nodes if two stocks belong to the same industry.
\ding{178} \textbf{DMFM} \cite{wei2022factor}  is a deep multifactor model built on top of a stock graph.
\ding{179} \textbf{VAE} \cite{duan2022factorvae} is a stock prediction method based on VAE that treats latent stock variables as the optimal factors thus learned.
\ding{180} \textbf{CVAE} \cite{gu2021autoencoder} is a factor model with a conditional autoencoder.

\subsection{Exp 1: Effectiveness of HireVAE}
In this experiment, we verify the effectiveness of our HireVAE with respect to factor investing.
In this context, factor models are the essential tools in factor investing, where the objective is to answer two practical questions: 
1) whether the model explains (future) stock returns well and 2) whether it provides good stock selection.

\subsubsection{Predictive Power}
To evaluate the predictive power of the compared methods, we use two metrics widely used in industry and academia \cite{duan2022factorvae,wei2022factor}, \emph{i.e.}, the rank information coefficient (IC), the rank information coefficient (RankIC), and the information ratio of RankIC (RankICIR) , where RankICIR is the z-score of RankIC. To reconcile deterministic and generative methods in prediction, we use the means of the hierarchical priors as the deterministic values and determine
the predicted stock return by $\hat{\mathbf{y}} = \phi_{dec}\left( \boldsymbol{\mu}_0^{s},\boldsymbol{\mu}_0^{m}, \mathbf{e}_s; c \right)$. 
By comparison in Table \ref{tab:pred-power}, HireVAE outperforms the compared methods, illustrating its effectiveness in stock prediction. 

\subsubsection{Stock Selection}
The most widely used method to test the ability of a factor model in portfolio construction is \emph{Top-1-of-G Groups}. Specifically, this involves sorting stocks in descending order \emph{w.r.t} the expected returns estimated by a factor model and then dividing the sorted stocks into $G$ groups. In addition, a long-short portfolio can be created by buying the \emph{Top-1} group and short selling the \emph{Bottom-1} group simultaneously. Since short selling in Chinese stock markets is not always possible for each stock, we construct only the portfolio holding the \emph{Top-1} group. The frequency of rebalancing is monthly. All transaction costs and taxes are included in our backtesting.

The CSI300, the CSI500, the CSI1000 and the CNI2000 are widely used benchmarks in China stock markets and consist of the most representative stocks in  market hierarchy. In particular, the CSI300 consists of the 300 most liquid and largest stocks. The CSI500, representing mid-caps, consists of the 500 most liquid and largest stocks. The CSI1000 consists of the following 1000 stocks representing small caps. In addition, the CNI2000 is another representative of small caps and consists of 2000 stocks.

To test the potential of subsequent use in the construction of enhanced index funds (EIFs), we compare the different methods. We report two metrics that are widely used in investment practice: the annualized active return (AR) relative to an investment benchmark and the information ratio of a portfolio (IR), the latter being a risk-adjusted measure. We use $\uparrow$ to indicate that a larger value is better. As shown in Table \ref{tab:portfolio}, the HireVAE can achieve better active returns over a long investment period across different market hierarchies.
\begin{table}[t]
    \centering
    \begin{tabular}{lrcrrrrrr}
        \toprule
        Method  &  Mkt &    rIC  &  rICIR & AR & IR  \\
        \midrule
        GRU     & \ding{51} &   0.051  & 0.484     & 13.180       & 2.202   \\
        GRU-s    & \ding{55} &   0.050   & 0.485    & 13.249       & 2.107   \\
        \midrule
        MLP     & \ding{51} &   0.051  & 0.537     & 19.247       & 3.525   \\
        MLP-s    & \ding{55} &   0.051  & 0.540     & 19.251       & 3.454   \\
        \midrule
        VAE     & \ding{51} &   0.059  & 0.539     & 23.724       & 4.377   \\
        VAE-s    & \ding{55} &   0.059  & 0.540     & 23.589       & 4.267 \\  
        \bottomrule
    \end{tabular}
    \caption{Ablation study of the difference in the use of market information. RIC rICIR, AR and IR are abbreviations for rank IC, rankICIR and active return, portfolio information ratio, respectively. The ``-s'' stands for the exclusive use of stock characteristics.}
    \label{tab:ablation-ms}
   \vspace{-0.3cm}
\end{table}

\subsection{Exp 2: Improvement Not From Extra Info} \label{market-info-the-same}
Theoretically, see our experiment settings in Section \ref{exp-setting}, we use all stocks included in the CSI All A Share Index, an overall market representative of Chinese stock markets. At its core, the index is roughly calculated by a capitalization-weighted sum of the prices of all stocks, which means that this index is calculated based on the most basic stock characteristics, price and market capitalization. Some of the original stock characteristics are also calculated based on these two basic characteristics. Thus, the market index is a ``global observation" of all stocks and no additional information is added to the original raw data.

To provide empirical support for the previous argument, we conduct two experiments.
First, all methods use the same input characteristics for stocks and the market, and the corresponding results are reported in Table\ref{tab:pred-power} and \ref{tab:portfolio}. Second, in Table \ref{tab:ablation-ms}, we show the comparisons between the baseline methods in terms of using market information, selecting the baseline methods that performed well in Exp 1 as our baseline methods in Exp 2. The results show that the additional market information does not help to improve the performance of a model.

\subsection{Exp 3: Adaptive Learning is Better}
Determining the regime on the basis of volatility is a way of establishing regimes on the basis of rules of thumb \cite{botte2021machine}. Inspired by this knowledge and experience, we develop a rule-based method shown in Algorithm \ref{alg:rule-based}. We consider this method as a baseline for regime switching to its online and adaptive counterpart. 
Table \ref{tab:ablation-regime} shows the performance of the different regime-switching algorithms in stock prediction and portfolio construction, where we select the most recommended stocks from the entire market (10\% stocks from the market). In different modeling settings, our adaptive regime switching algorithm can always outperform the rule-based and neural clustering counterparts.

\begin{table}[t]
    \centering
    \begin{tabular}{lrcrrrrrr}
        \toprule
        Method  &  Algo &    Enc  &  rIC & rICIR & AR  \\
        \midrule
        RVAE-v    & Rule &   \ding{55}  & 0.05     & 0.67       & 21.62   \\
        RVAE     & Clus &  \ding{55}  & 0.05          & 0.50       & 13.72   \\
        RCVAE-v  & Rule &  \ding{55}    & 0.04          & 0.38       & 14.85  \\
        RCVAE    &  Clus & \ding{55}& 0.03          & 0.32       & 2.61   \\
        HiReVAE-v  &  Rule &\ding{51}     & 0.06          & 0.67       & 24.54    \\
        \textbf{HiReVAE}   & \textbf{AOL} & \ding{51}       & \textbf{0.07}         & \textbf{0.73}       & \textbf{28.02}   \\
        \bottomrule
    \end{tabular}
    \caption{Ablation study on the rule-based and the end-to-end regime switching identifier in terms of their predictive power and stock selection performance on all Chinese A-shares. The rIC, rICIR, AR are abbreviations for rank IC, rankICIR and active return, respectively. ``-v'' stands for regime change based on rule-based algorithm.}
    \label{tab:ablation-regime}
\end{table}

\begin{algorithm}[tb]
    \caption{Volatility-based regime identification}  
    \label{alg:rule-based}
    \textbf{Input}: Volatility data $\mathcal{D}_{\sigma}= \{ \sigma_{t_0}, \dots, \sigma_{t_n}\}$ where all volatility data items are sorted in chronological order, and the number of regimes $K$ ($K \geq 2$) \\
    \textbf{Output}: Regime ${c}_t$
    \begin{algorithmic}[1] 
        \STATE Initialize $\sigma_{min} = +\infty$ and $\sigma_{max} = -\infty$
        \STATE Initialize $I_K = \left[\frac{(K-1)}{K}, 1 \right]$ and $I_k = \left[\frac{(k-1)}{K}, \frac{k}{K}\right)$, for $ k = 1, \dots, K-1$
        \WHILE{$ t \leq t_n$ }
        \STATE $\sigma_{min} \leftarrow \min(\sigma_{min}, \sigma_t)$ 
        \STATE $\sigma_{max} \leftarrow \max(\sigma_{max}, \sigma_t)$ 
        \STATE $\sigma_{level} \leftarrow (\sigma_t-\sigma_{min})/({\sigma_{max}-\sigma_{min}})$ 
        \WHILE{$ k \leq K$}
        \IF {$\sigma_{level} \in I_k$}
        \STATE ${c}_t \leftarrow k$.
        \ENDIF
        \ENDWHILE
        \ENDWHILE
        \STATE \textbf{return} regime ${c}_t$
    \end{algorithmic}
\end{algorithm}

\subsection{Exp 4: Hierarchy is Necessary}
In this experiment, we analyze the necessity of using a hierarchical structure in two ways: first, does a hierarchical latent space (HVAE) encode market information better than a single latent space (VAE)? Second, does the hierarchical structure help the end-to-end clustering algorithm in adaptive learning of market regimes.

Table \ref{tab:ablation-hirevae} can answer these two questions accordingly. As for the first question, encoding market information alone does not provide significant improvement in the non-regime-swithcing modeling, since the metrics of VAE and HiVAE are similar. This is reasonable, as we showed in Section \ref{market-info-the-same}, where the market information itself does not provide more information than the raw data. However, in the case of regime switching, the market latent variable plays an important role in end-to-end regime switching learning, as it is used to estimate the regime region and predict the corresponding indicators that measure the probability of being in each regime. This can be empirically demonstrated by the fact that HireVAE outperforms the other two regime switching methods without considering the market latent variable, RVAE and RCVAE, in terms of predictive power and stock selection ability.

\begin{table}
    \centering
    \begin{tabular}{lrrrrr}
        \toprule
    Method  & Hier & MLV & rIC & rICIR & AR  \\
        \midrule
      VAE     & \ding{55} & \ding{51}& 0.06 & 0.63 & 23.72  \\
     HiVAE   & \ding{51} & \ding{51}& 0.07 & 0.68 & 23.13  \\
    \midrule
      RVAE    & \ding{55} & \ding{51}&  0.05 & 0.50 & 13.72\\
    RCVAE   & \ding{55} & \ding{51}& 0.03 & 0.32 & 2.61   \\
      \textbf{HiReVAE}  & \ding{51} & \ding{51}& \textbf{0.07} & \textbf{0.73} & \textbf{28.02}  \\
    \bottomrule
    \end{tabular}
    \caption{The result of the experiment showing the need for a hierarchical structure in learning an online and adaptive regime switching identifier. The MLV is the abbreviation for the market latent variable, which indicates whether the MLV participates in regime switching learning.}
    \label{tab:ablation-hirevae}
\end{table}

\section{Conclusion}
In this paper, we present HireVAE, a novel end-to-end neural factor model that can identify current market regime according to point-in-time market information, and subsequently adapt itself for better prediction.
HireVAE achieves such an ability by a pair of hierarchically organized encoders respectively process global market situation and stock-wise latent factors. 
Being the first online and adaptive regime-switch factor model, HireVAE achieves superior performance on real stock market data, outperforming classical linear methods and recent non-linear data-driven methods by a large margin.

\appendix


\section*{Acknowledgments}

This project is funded in part by Shanghai AI Laboratory, 
CUHK Interdisciplinary AI Research Institute, 
and the Centre for Perceptual and Interactive Intelligence (CPII) Ltd 
under the Innovation and Technology Commission (ITC)'s InnoHK.

\bibliographystyle{named}
\bibliography{ijcai23}

\end{document}